\pdfoutput=1

\documentclass[11pt]{article}
\usepackage[final]{coling}

\usepackage{times}
\usepackage{latexsym}
\usepackage[T1]{fontenc}
\usepackage[utf8]{inputenc}
\usepackage{microtype}
\usepackage{inconsolata}
\usepackage{graphicx}
\usepackage{booktabs}
\usepackage{adjustbox}
\usepackage{xcolor,colortbl}
\usepackage{enumitem}
\usepackage{array}
\usepackage{listings}
\usepackage{adjustbox}
\usepackage{wrapfig}
\usepackage{amsmath}
\usepackage{amsfonts}
\usepackage{tcolorbox}
\usepackage{multicol}
\usepackage[T1]{fontenc}
\usepackage[utf8]{inputenc}
\usepackage{microtype}
\usepackage{inconsolata}

\title{In-context Continual Learning Assisted by an External Continual Learner}
\author{
    Saleh Momeni~\textsuperscript{1},~ 
    Sahisnu Mazumder~\textsuperscript{2},~ 
    Zixuan Ke~\textsuperscript{3},~ 
    Bing Liu~\textsuperscript{1}
    \\
    \textsuperscript{1}~Department of Computer Science, University of Illinois Chicago, USA\\
    \textsuperscript{2}~Intel Labs, USA~~~~
    \textsuperscript{3}~Salesforce AI Research, USA\\
    \texttt{smomen3@uic.edu,~sahisnumazumder@gmail.com,} \\
    \texttt{zixuan.ke@salesforce.com,~liub@uic.edu} \\
}

\begin{document}
\maketitle
\begin{abstract}
Existing continual learning (CL) methods mainly rely on fine-tuning or adapting large language models (LLMs). They still suffer from \textit{catastrophic forgetting} (CF). Little work has been done to exploit in-context learning (ICL) to leverage the extensive knowledge within LLMs for CL \textit{without} updating any parameters. However, incrementally learning each new task in ICL necessitates adding training examples from each class of the task to the prompt, which hampers scalability as the prompt length increases. This issue not only leads to excessively long prompts that exceed the input token limit of the underlying LLM but also degrades the model’s performance due to the overextended context. To address this, we introduce InCA, a novel approach that integrates an \textit{external continual learner} (ECL) with ICL to enable scalable CL without CF. The ECL is built incrementally to pre-select a small subset of likely classes for each test instance. By restricting the ICL prompt to only these selected classes, InCA prevents prompt lengths from becoming excessively long, while maintaining high performance. Experimental results demonstrate that InCA significantly outperforms existing CL baselines, achieving substantial performance gains.\footnote{Published as a conference paper at COLING 2025.}

\end{abstract}

\section{Introduction}
\label{sec:introduction}
Continual learning (CL) aims to enable models to learn a sequence of tasks incrementally \citep{chen2018lifelong,de2021continual}. CL is typically categorized into three main settings: \textit{task-incremental learning}, \textit{class-incremental learning} (CIL), and \textit{domain-incremental learning} \cite{van2019three}. In this paper, we focus on the CIL setting \citep{rebuffi2017icarl}, where each task has a set of distinctive classes, and a single model is developed to handle all tasks and classes. At test time, no task information is provided for each test instance. This differs from task-incremental learning, which provides the task-id for each test instance, making classification much easier. CIL requires a unified model that can distinguish all classes seen thus far. In domain-incremental learning, all tasks have the same classes but are from different domains.  

There are two key challenges in CIL. \textbf{(1) \textit{Catastrophic forgetting}} (CF), which refers to the performance deterioration of earlier tasks due to parameter updates in learning new tasks \citep{McCloskey1989}. \textbf{(2) \textit{Inter-task class separation}} (ICS), which refers to the phenomenon that without accessing the previous task data, the learning of a new task has difficulty in establishing decision boundaries between the new and old classes \citep{kim2023learnability}. Although the CL community has studied CF extensively, the challenge of ICS has only been identified recently in \citep{kimtheoretical}. Both challenges disappear in \textit{\textbf{in-context CIL}} with LLMs. A simple method to apply in-context learning to CIL is to incrementally add few-shot training examples for each new class to the in-context prompt. This prompt includes examples from all classes encountered so far along with instructions for classification. Since the LLM parameters remain unchanged, CF is inherently avoided, and ICS is addressed by encompassing all classes and their examples within the same prompt.

Unfortunately, this approach is not scalable for CIL because the prompt length rapidly increases with each new task or class added, quickly exceeding the token limits of LLMs. Although summarizing the training examples can increase the number of classes that can be learned (i.e., included in the prompt), the underlying scalability problem persists. Moreover, including excessive and often irrelevant information from various classes leads to significant performance degradation (see Section \ref{sec.results}). Even with the recently introduced long-context LLMs \citep{chen2023longlora, reid2024gemini}, our experiments demonstrate that the performance degradation persists despite the increased token capacity.

\begin{figure*}[t]
    \centering
    \includegraphics[width=\textwidth]{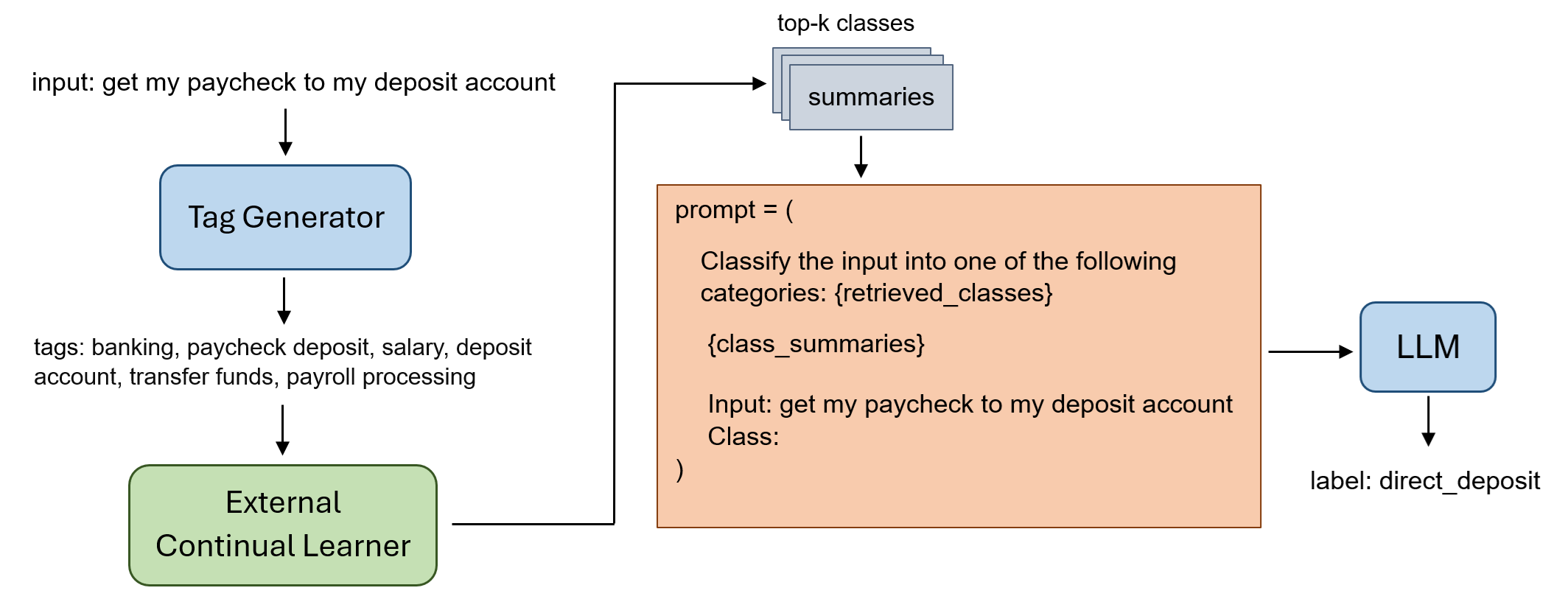}
    \caption{Overview of the InCA framework. The diagram depicts the stages of generating semantic tags for the input, identifying the most similar classes via the ECL, and constructing the prediction prompt with class summaries, which together enables efficient in-context continual learning without retaining any training data.}
    \label{pipeline}
\end{figure*}

In this paper, we introduce InCA (\textbf{In}-context \textbf{C}ontinual Learning \textbf{A}ssisted by an External Continual Learner), a novel method that overcomes the scalability and performance limitations of in-context CIL while retaining the advantages of in-context learning -- specifically, avoiding CF and ICS problems. InCA leverages an \textit{external continual learner} (ECL) that both benefits from and enhances the LLM's in-context learning capabilities.
The ECL aims to reduce the number of candidate classes to a small set of $k$ classes that are most likely to include the correct class. For each input instance, we first prompt the LLM to generate a list of \textit{tags}--descriptive topics or keywords that capture the essential semantics of the input text (see Figure \ref{pipeline} for an illustrative example). Each class in the dataset is represented by a Gaussian distribution over the embeddings of these tags, characterized by a \textit{mean vector} and a \textit{shared covariance matrix}. 
The ECL then computes the Mahalanobis distance \citep{de2000mahalanobis} between the input’s tag embeddings and each class distribution to identify the top \textit{k} most similar classes. These selected classes are then used to construct an in-context learning prompt, efficiently managing the token limit while removing irrelevant information.\footnote{Our experiments demonstrate that the ECL achieves high top-\textit{k} recall, ensuring that the correct class is almost always included in the top $k$ classes to be used in the final in-context learning prompt.}

Unlike traditional CL methods, our ECL requires no additional training -- it only incrementally accumulates and updates class means and a shared covariance matrix derived from the embeddings of the tags generated by the LLM. This approach inherently avoids CF. Moreover, representing each class with a Gaussian distribution addresses the ICS problem, as different classes are naturally distinguished by their statistical distributions.
While the ECL alone (e.g., performing top-1 classification based on Mahalanobis distance) can be applied to CIL, its standalone accuracy is limited. However, when integrated with the LLM's in-context learning, InCA significantly improves performance, as demonstrated in our experiments (see Section \ref{sec.results}). This approach effectively balances scalability and accuracy, making in-context CIL feasible and efficient.

To summarize, our contributions are as follows:
\begin{enumerate}
\item We introduce the novel paradigm of in-context CIL, which, to the best of our knowledge, has not been previously studied.
\vspace{-1pt}
\item We propose InCA, a new method that addresses token limit constraints and performance degradation caused by overextended context in in-context CIL.
\vspace{-1pt}
\item Our method surpasses existing state-of-the-art CIL baselines, achieving significant performance improvements across different benchmark datasets.
\end{enumerate}

\section{Related Works}
\label{sec:related works}
There is a large body of literature on continual learning. The main focus is on dealing with CF.
Existing techniques can be broadly classified into a few categories. (1) \textit{Regularization}, which uses a regularizer to ensure that important network parameters from previous tasks are minimally altered when learning new tasks, thereby reducing CF \citep{li2022overcoming,liu2019continual}. (2) \textit{Replay}, which stores some training samples from previous tasks. When learning a new task, the model is trained using both the new task data and the stored replay data to mitigate CF \cite{liu2021lifelong,qin2022elle,huang2021continual}. Some replay methods do not store actual data but learn data generators to generate data similar to those from previous tasks \citep{shin2017continual,he2018overcoming}. (3) \textit{Architectural-based}, which encompasses various methods aimed at managing CF through structural modifications. Some techniques expand the network's capacity as new tasks are learned \cite{wang2022beef,yan2021dynamically, qin-etal-2023-lifelong}. Some do \textit{parameter isolation}, which trains sub-networks for each task by using masks to prevent updates to critical parameters or neurons from previous tasks, or by ensuring that new task parameters are orthogonal to those of prior tasks \cite{ke2021achieving,ke2023continual,konishi2023parameter,Serra2018overcoming,gururangan2021demix,zhu2022continual,geng2021continual,lin2022beyond, liu2023continual}. 

Moreover, some methods incorporate parameter-efficient fine-tuning (PEFT) techniques such as low-rank adaptation (LoRA) \cite{hu2021lora} and prompt-tuning to allocate task-specific parameters for each new task \citep{razdaibiedina2023progressive, wang2022learning, wang2024rehearsal}.
These systems often implement various mechanisms to predict the task-id, which is essential for selecting the appropriate model for CIL. They may utilize a separate network, entropy, or out-of-distribution detection to predict the task-id \citep{rajasegaran2020itaml,abati2020conditional,kim2023learnability}. Our work differs from these approaches as it does not require task-id prediction. While the aforementioned methods train a different model for each task and rely on task-ids, our approach allows for adding one class at a time with no task-id prediction. Our ECL directly predicts the most probable classes.

In the field of NLP, CL has been employed to address a variety of problems, including text classification \citep{chuang2020lifelong}, sentiment analysis \citep{ke2021achieving}, topic modeling \citep{gupta2020neural}, slot filling \citep{shen2019progressive}, question answering \citep{greco2019psycholinguistics}, language learning \citep{li2019compositional, liang2024continual, zhao2024sapt}, and the pre-training of language models \cite{ke2023continual, qin2022elle}. Pre-trained models are commonly utilized in most NLP-related continual learning tasks, serving as a standard practice \citep{ke2021adapting, wang-etal-2024-inscl}. For further insights and a comprehensive overview, refer to surveys \citep{ke2022continualsurvey,wang2023comprehensive}.

Our approach differs from the aforementioned methods that adapt or fine-tune pre-trained language models, as we primarily leverage in-context learning for CIL. While advancements in LLMs have improved few-shot and instruction-based prompting \cite{wei2022chain, yao2023tree, hao2023reasoning}, they fail to address CL challenges, such as growing prompt sizes that quickly exceed token limits. Moreover, even with long-context LLMs \citep{reid2024gemini, dubey2024llama}, extended prompts containing excessive and often irrelevant information can lead to significant performance degradation. InCA overcomes these issues by using an external continual learner that can be updated without CF.

InCA bears some resemblance to retrieval-augmented generation (RAG) \cite{zhao2024retrieval}, which  uses a retriever to gather information to provide domain-specific knowledge for enhancing content generation. However, InCA is fundamentally different, as our ECL is a coarse-grained classifier that tries to identify the most similar classes rather than retrieving domain- or task-specific content. Additionally, due to the incremental nature of CIL, our ECL must be built incrementally and handle CF and ICS without storing data from previous tasks -- challenges that retrievers do not encounter.

\section{Problem Formulation}
We study class-incremental learning in the text classification domain. CIL involves learning a sequence of tasks arriving sequentially \cite{kim2023learnability}. Let $B$ be the number of tasks encountered so far. Each task $b$ ($1 \leq b \leq B$) is associated with a training dataset $\mathcal{D}_b = \{(x_b^{(i)}, y_b^{(i)})\}_{i=1}^{n_b}$, where $n_b$ is the total number of instances in $\mathcal{D}_b$, $x_b^{(i)}$ denotes an input (text) instance, and $y_b^{(i)}$ is its corresponding class label. Let $\mathbf{Y}_b$ be the set of classes belonging to $b$ (i.e., the set of all classes in $\mathcal{D}_b$). For any two tasks $b$ and $b'$, their corresponding class sets are disjoint ($\mathbf{Y}_b \cap \mathbf{Y}_{b'} = \emptyset$ for $b \neq b'$). The overall class set for all $B$ tasks is defined as $\bigcup_{b=1}^B \mathbf{Y}_b = \mathbf{Y}$. The goal is to construct a unified predictive function $f : \mathbf{X} \rightarrow \mathbf{Y}$ capable of classifying any given test instance $x$ across all tasks/classes seen so far, despite the restriction that no data from previous tasks are retained during training, i.e., \textbf{replay-free}.

\section{Proposed InCA Method}
\label{sec:rcl}
This section presents InCA (In-context Continual Learning Assisted by an External Continual Learner), a framework designed to address the challenges of CIL by leveraging the in-context learning capabilities of LLMs. InCA has three main stages: (1) \textbf{tag generation}, where semantic tags are extracted from the input text using the LLM (Section \ref{sec:tag-generation}); (2) \textbf{external continual learning}, which identifies the top $k$ most probable classes based on the generated tags through Gaussian class modeling and Mahalanobis distance scoring (Section \ref{sec:ECL}); and (3) \textbf{in-context learning with class summaries}, where the LLM predicts the final class label for the input text using summaries of the top $k$ candidate classes (Section \ref{sec:summaries}). Figure \ref{pipeline} depicts the overall framework.

\subsection{Tag Generation}
\label{sec:tag-generation}
To capture the essential semantic information from an input text $x$, we generate a list of tags that include topics, keywords, important entities, and other relevant elements. For example, a customer’s banking query processed by our framework (see Figure \ref{pipeline}) might generate tags such as ``banking'' or ``paycheck deposit'' while omitting less pertinent information. Additionally, the tags are automatically extended to include related terms that commonly appear in similar contexts. For instance, the tag ``paycheck deposit'' may be extended to include terms like ``transfer funds'' or ``payroll processing''. Tags are generated by prompting the LLM to produce both primary tags and related terms. The specific prompt used for tag generation is detailed in Appendix~\ref{tag-generation-prompt}.

\subsection{External Continual Learner}
\label{sec:ECL}
The ECL leverages the generated tags to identify the $k$ most probable classes for a given input, thereby filtering out the irrelevant context. As mentioned earlier, the ECL operates by accumulating statistics without additional training and thus, inherently avoids CF.

\textbf{Gaussian Class Representation}: Each class is modeled as a Gaussian distribution, with a mean vector and a shared covariance matrix. This representation helps mitigate the ICS problem by allowing classes to have independent distributions. However, since the covariance matrix has high dimensionality, storing a separate covariance matrix for each class would result in excessive space consumption. 
To address this, we assume that all classes share the same covariance matrix, drastically reducing the space required.

Let \( \mathcal{T}_j = [ t_{1,j}, t_{2,j} \dots, t_{R,j} ] \) be the list of all tags generated by the LLM for class \( j \), where \( R \) is the total number of tags generated from all training instances of class \( j \). We employ the widely-used Sentence-BERT (SBERT) \citep{reimers2019sentence} model to encode each tag $t_{r,j} \in \mathcal{T}_j$ into a \( h \)-dimensional embedding vector \( z_{r,j} \in \mathbb{R}^h \). The mean vector \( \mu_j \in \mathbb{R}^h \) for class \( j \) is computed as the average of all its tag embeddings:
\[
\mu_j = \frac{1}{R} \sum_{r=1}^{R} z_{r,j}
\]

The shared covariance matrix \( \Sigma \in \mathbb{R}^{h \times h} \) is updated incrementally as new classes are introduced. The contribution of class $j$ to the shared covariance matrix, denoted as $\Delta_j$, is based on the deviations of its tag embeddings from the mean \citep{park2018fundamentals}:
\[
\Delta_j = \frac{1}{R} \sum_{r=1}^{R} (z_{r,j} - \mu_j)(z_{r,j} - \mu_j)^T
\]

The overall shared covariance matrix is updated after each new class is processed:
\[
\Sigma_{j} = \frac{(j-1) \Sigma_{j-1} + \Delta_j}{j},
\]
where \( \Sigma_j \) denotes the shared covariance matrix after processing class \( j \), assuming that classes \( \{1, 2, \dots, j-1\} \) have been previously learned.

\vspace{1mm}
\textbf{Mahalanobis Distance Scoring}: For each test instance, the ECL uses the Mahalanobis distance \citep{de2000mahalanobis} to select the top $k$ most similar classes. Let \( \{z_i\}_{i=1}^{m} \) be the set of tag embeddings generated for the input instance $x$. The Mahalanobis distance between an embedding \( z_i \) and the Gaussian distribution for class \(j\) is computed as:
\[
d(z_i, \mu_j, \Sigma) = \sqrt{(z_i - \mu_j)^T \Sigma^{-1} (z_i - \mu_j)}
\]
Here, $\Sigma$ represents the shared covariance matrix updated up to the current point when the inference is performed. The overall distance of the test instance \( x \) from the class is the average Mahalanobis distance over all tag embeddings:
\[
d(x, \mu_j, \Sigma) = \frac{1}{m} \sum_{i=1}^{m} d(z_i, \mu_j, \Sigma)
\]
The top $k$ classes with the smallest Mahalanobis distances are selected for the final prediction step.

\subsection{In-context Learning with Class Summaries}
\label{sec:summaries}
Once the top $k$ candidate classes are identified by the ECL, in-context learning is applied using \textbf{class summaries} to determine the final prediction. Storing and using too many training examples for in-context learning would be impractical and inefficient for continual learning. Instead, for each class, we generate a summary at the time of its introduction, serving as a compact representation of the class.

\vspace{1mm}
\textbf{Generating Class Summaries}: The summary for each class is generated by prompting the LLM using a small subset of randomly selected training examples of the class. The prompt used for generating the summary is given in Appendix~\ref{summarization-prompt}. Each summary captures the essential characteristics or information of the class, allowing for efficient in-context learning without storing lots of examples.

\textbf{Prediction with In-context Learning}: During prediction, the test instance is concatenated with the summaries of the top \( k \) classes to form a single prompt. The LLM processes this prompt and predicts the class label based on the context provided. The prompt format for this prediction stage is detailed in Appendix~\ref{prediction-prompt}.

\vspace{1mm}
To summarize, InCA stores only a mean embedding vector for each class and a shared covariance matrix for all classes encountered so far. Thus, the amount of information saved in the CIL process is very small. The whole process involves \textbf{no training} and it is \textbf{replay-free}, i.e., no previous task data is stored to help deal with CF. Moreover, as explained earlier, it avoids both the CF and ICS problems that have plagued the existing CIL techniques.

\section{Experiment Setup}
In this section, we describe the datasets, baselines, implementation details, and evaluation metrics.

\textbf{Datasets:} We utilize four datasets for our experiments: CLINC \citep{larson2019evaluation}, Banking \citep{casanueva2020efficient}, HWU \citep{liu2021benchmarking}, and DBpedia \citep{auer2007dbpedia}. The intent classification datasets -- CLINC, Banking, and HWU comprise of 150, 77, and 64 classes, respectively. DBpedia is a topic classification dataset with 70 classes. For the train/test splits, we allocate 10k/750 samples for CLINC, 10k/1k samples for Banking, 9k/1k samples for HWU, and 10k/1k samples for DBpedia.\footnote{Note that we do not use datasets for some other NLP tasks (e.g., dialogue generation, summarization, translation, etc.) because they are not suitable for class-incremental learning. More details can be found in Section \ref{sec.limitations}.}
InCA can incrementally learn each class one by one. For baselines, we adhere to the standard CIL protocol by splitting the classes into disjoint tasks, each composed of a subset of classes. Multiple runs with different task splits are conducted, and the accuracy values are averaged to minimize the influence of any specific task split on overall performance.

\begin{table*}[t]
\centering
\scalebox{0.9}{
\begin{tabular}{llccccccc}
\toprule
\multicolumn{2}{c}{} & \multicolumn{5}{c}{Fine-tuning based Methods} & \multicolumn{2}{c}{} \\
\cmidrule(lr){3-7}
Dataset & \#Tasks & Vanilla & EWC & L2P & LAMOL & VAG & InCA & \cellcolor[gray]{0.8}JOINT \\
\midrule
CLINC & 10 & $51.27_{\scriptsize\ \pm 1.26}$ & $54.22_{\scriptsize\ \pm 1.14}$ & $52.53_{\scriptsize\ \pm 1.72}$ & $58.42_{\scriptsize\ \pm 0.84}$ & $76.42_{\scriptsize\ \pm 0.90}$ & \textbf{94.40} & \cellcolor[gray]{0.8}97.60 \\
Banking & 7 & $27.77_{\scriptsize\ \pm 2.46}$ & $29.10_{\scriptsize\ \pm 1.78}$ & $25.78_{\scriptsize\ \pm 1.21}$ & $42.60_{\scriptsize\ \pm 1.36}$ & $59.34_{\scriptsize\ \pm 1.28}$ & \textbf{84.90} & \cellcolor[gray]{0.8}92.50 \\
DBpedia & 7 & $39.02_{\scriptsize\ \pm 2.68}$ & $40.30_{\scriptsize\ \pm 2.89}$ & $42.84_{\scriptsize\ \pm 5.47}$ & $48.61_{\scriptsize\ \pm 1.82}$ & $65.40_{\scriptsize\ \pm 1.52}$ & \textbf{84.20} & \cellcolor[gray]{0.8}95.70 \\
HWU & 8 & $38.38_{\scriptsize\ \pm 4.01}$ & $42.72_{\scriptsize\ \pm 2.62}$ & $28.77_{\scriptsize\ \pm 3.18}$ & $44.85_{\scriptsize\ \pm 1.57}$ & $56.88_{\scriptsize\ \pm 1.22}$ & \textbf{86.61} & \cellcolor[gray]{0.8}90.43 \\
\bottomrule
\end{tabular}
}
\caption{Final accuracy (\%) of InCA compared with various fine-tuning based baselines. The gray column shows the results in the JOINT setting, which is not continual learning and regarded as the \textbf{upper bound}. Experiments are conducted with three different task splits to minimize the influence of any specific task split on performance.}
\label{fine-tuning-results}
\vspace{-5pt}
\end{table*}

\textbf{Baselines:} We evaluate our proposed method against several baselines. The \textbf{Vanilla} baseline sequentially fine-tunes the LLM on each task without any specific mechanism to mitigate CF. \textbf{EWC} \citep{kirkpatrick2017overcoming} is a regularization-based technique that mitigates CF by preserving important parameters for previous tasks through a quadratic penalty. \textbf{L2P} \citep{wang2022learning} freezes the LLM's parameters and learns a set of trainable prompts to guide the model during inference. \textbf{LAMOL} \citep{sun2020lamol} employs pseudo-replay, generating pseudo-examples of previous tasks to interleave with new task data in training, thus maintaining performance on past tasks. \textbf{VAG} \citep{shao2023class} leverages vocabulary sparsity to selectively generate relevant outputs for each task through label generation, rather than traditional classification objective. \textbf{JOINT} learns all the classes together as a single task. It is not a continual learning setting and its accuracy is regarded as the upper-bound accuracy of CIL.

Additionally, we compare our method against several \textbf{long-context LLMs}, where all class summaries are directly included in the final in-context learning prompt (see Section \ref{sec:summaries}) \textit{without} using the ECL. Specifically, we evaluate using Mistral (with a 32K context window) \citep{jiang2023mistral}, Llama3 (128K) \citep{dubey2024llama}, and Gemini (2M) \citep{reid2024gemini}, each trained to support these extended context lengths. We also include comparisons with LongLlama (128K) \citep{tworkowski2024focused} and LongAlpaca (16K) \citep{chen2023longlora}, which adapt pre-trained LLMs for handling longer contexts.

\textbf{Implementation Details:} We use the \textbf{Mistral-7B} model for all experiments and ablations, except for one where we evaluate our framework with different LLMs, specified accordingly. For generating summaries, we use 20 training instances per class in the summarizing prompt. \footnote{We also tested with more than 20 instances per class for summaries but observed no noticeable improvement.} In the ablation with limited data, we use all available instances when there are fewer than 20. We limit the length of the summaries to a maximum of 256 tokens. To determine the optimal $k$ value for the ECL, we use a small validation set. Despite varying values for $k$ across different datasets, it never exceeds 3 in any dataset during our experiments, indicating that including only 3 class summaries in the final prompt is sufficient for accurate prediction, even with 150 classes as in the CLINC dataset.

For the Vanilla, EWC, and JOINT systems, we perform parameter-efficient fine-tuning using LoRA adaptors \citep{hu2021lora}. This approach is selected due to the large model size and PEFT's better adaptation to the task with limited data. Following \citep{shao2023class}, we employ a label generation objective instead of a classification head, as generation loss helps mitigate CF. For LAMOL and VAG, we had to use their original language models, BART for VAG and GPT-2 for LAMOL, because LAMOL's code is incompatible with Mistral, while VAG's use of an encoder-decoder architecture also results in incompatibility. For embedding the tags, we use a small SBERT paraphrase-MiniLM-L6-v2 model. All pre-trained models are obtained from the Transformers library \citep{wolf2020transformers}.

Experiments were conducted on a single A100 GPU with 80GB VRAM. The baseline models implemented were trained for 10 epochs per task with a batch size of 8, employing early stopping and the Adam optimizer with a learning rate of 2e-5. The LAMOL and VAG models were executed using their official configurations and hyperparameter settings. All models were maintained at 32-bit precision during training and inference, except for long-context settings, where 8-bit quantization \citep{dettmers2208llm} 
 was employed due to the extended prompt length and VRAM requirements.

\textbf{Evaluation Metric:} We measure classification accuracy after all tasks/classes have been processed, referred to as \textbf{Last or Final accuracy}. All experiments are conducted three times and the accuracy is averaged, except for the zero-training LLM setups, which are deterministic.

\section{Main Results}
\label{sec.results}
\textbf{Surpassing Traditional CIL:} The proposed InCA demonstrates a clear advantage over traditional CIL methods involving training, as seen in Table \ref{fine-tuning-results}. It significantly outperforms all the baselines across all datasets. Despite employing a range of strategies such as regularization, parameter freezing, and pseudo-replay, none of the baseline methods achieved comparable performance.

As outlined in Section \ref{sec:introduction}, CF occurs when updating model parameters for a new task disrupts the knowledge acquired from previously learned tasks. Our proposed system, InCA, avoids CF as it operates without any training. Although InCA's performance remains below the upper bound achieved by JOINT fine-tuning, this is mainly due to the limitations of in-context learning, which may not match the task-specific optimization of fine-tuning.

\begin{figure*}[t]
    \centering
    \includegraphics[width=\textwidth]{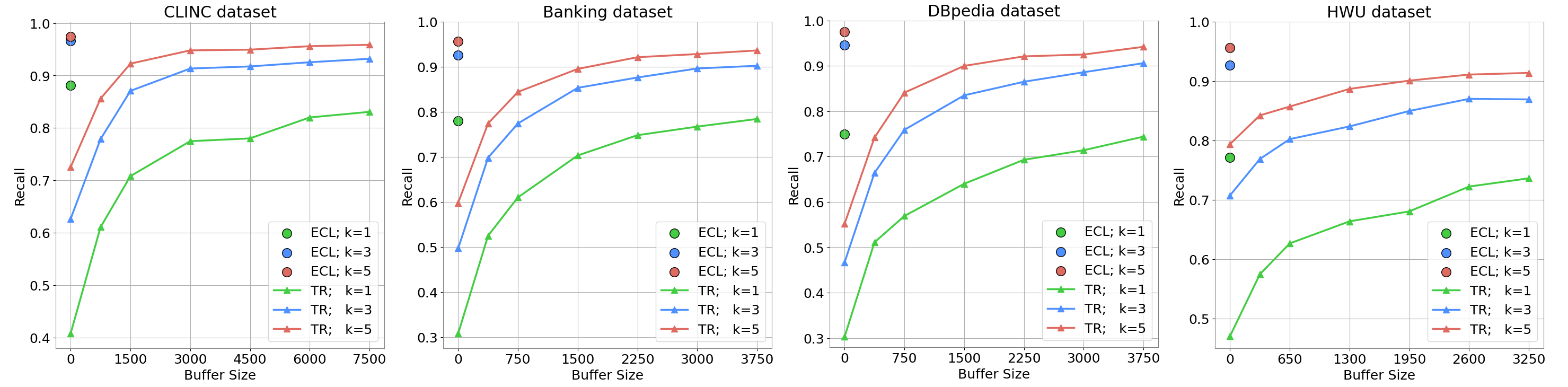}
    \caption{Comparison of recall between our ECL and the text retriever (TR) at various values of $k$. The ECL operates without storing any replay data (buffer size = 0), while the TR maintains a buffer of instances to retrieve the most similar ones during inference. For the TR, we evaluate performance across different buffer sizes. When the TR's buffer size is zero, we store the embeddings of class summaries for retrieval, rather than training instances.}
    \label{rag}
    \vspace{-5pt}
\end{figure*}

\textbf{InCA vs. Long-context Setting:} The limited context window of LLMs becomes a significant challenge as the number of classes increases. InCA addresses this by using an external continual learner to identify the most relevant classes and construct a precise prediction prompt. In contrast, extending the context window of the LLM to include more information is an alternative approach. To assess the effectiveness of our approach, we compared InCA against the long-context setting of LLMs, where all class summaries are passed directly into the prediction prompt.

We conducted experiments using several long-context models, including Mistral, Llama3, and Gemini 1.5 flash, both with and without the assistance of the ECL. Additionally, we tested LongLlama and LongAlpaca, which adapt standard LLMs to handle long-context tasks. Since these models are not instruction-tuned, they are not suitable for generating the tags required by InCA. Consequently, we used these models only in the long-context setting, utilizing class summaries generated by Mistral, as their own summaries may not match the quality of instruction-tuned models.

The results, shown in Table \ref{long-context-results}, reveal that the long-context models without the ECL performed significantly worse than InCA. Even with Gemini, which has a 2M token context window, performance degraded when overloaded with excessive context. This demonstrates that extending the context length is not sufficient, as overextended prompts hamper the model’s ability to focus on the relevant information. In contrast, InCA ensures that only the most relevant class information (summaries) is included in the prompt, resulting in more accurate predictions and also, faster inference times due to the shorter prompts.

\begin{table}[t]
\centering
\resizebox{\columnwidth}{!}{
\begin{tabular}{l|cccc}
\toprule
Model & CLINC & Banking & DBpedia & HWU \\
\hline
Mistral-7B   & 94.40\% & 84.90\% & 84.20\% & 86.61\% \\
Llama3-8B    & 95.73\% & 84.30\% & 87.60\% & 87.45\% \\
Gemini 1.5 flash    & 95.32\% & 86.15\% & 91.63\% & 89.22\% \\
\hline
& \multicolumn{4}{c}{Without ECL} \\
\hline
Mistral-7B     & 86.93\% & 65.90\% & 65.30\% & 81.04\% \\
Llama3-8B      & 83.73\% & 77.80\% & 72.70\% & 83.27\% \\
Gemini 1.5 flash      & 93.86\% & 83.52\% & 79.64\% & 87.27\% \\
LongAlpaca-7B   & 45.87\% & 33.20\% & 24.90\% & 35.97\% \\
LongAlpaca-13B  & 51.20\% & 63.60\% & 59.10\% & 62.83\% \\
LongLlama-3B   & 62.00\% & 52.80\% & 38.90\% & 58.46\% \\
LongLlama-7B   & 84.67\% & 73.10\% & 61.00\% & 77.88\% \\
\bottomrule
\end{tabular}
}
\caption{Comparison of InCA against long-context LLMs (without ECL), where all class summaries are included in the prediction prompt. For LongLlama and LongAlpaca models, class summaries are generated using Mistral, as they are not instruction-tuned.}
\label{long-context-results}
\vspace{-12pt}
\end{table}

\section{Ablation Results}
\vspace{-3pt}
\textbf{Tag-based ECL vs. Text Retrieval:} 
Given the resemblance between our approach and retrieval-augmented generation, we benchmark our tag-based classifier against a retrieval method based on text similarity. We adopt a common RAG framework, maintaining a pool of training instances for each class and using SBERT to retrieve the most similar instances during inference based on text similarity. It is important to note that this method is not applicable to CL since it involves storing original instances; we use it solely for comparison. We retrieve instances until we have obtained instances from $k$ distinct classes, after which we measure the recall and compare these results to our ECL, which operates without a buffer (i.e., no replay data).

The results illustrated in Figure \ref{rag} demonstrate that our tag-based ECL consistently outperforms text retrieval across all scenarios, even when the text retriever has access to a substantial buffer of training instances. The same SBERT model is used for embedding both tags and input text. The superior performance of the ECL is particularly noteworthy as it does not require storing any instances, highlighting its effectiveness, especially in contexts where storing training instances is impractical.

\begin{figure*}[h]
    \centering
    \includegraphics[width=0.77\textwidth]{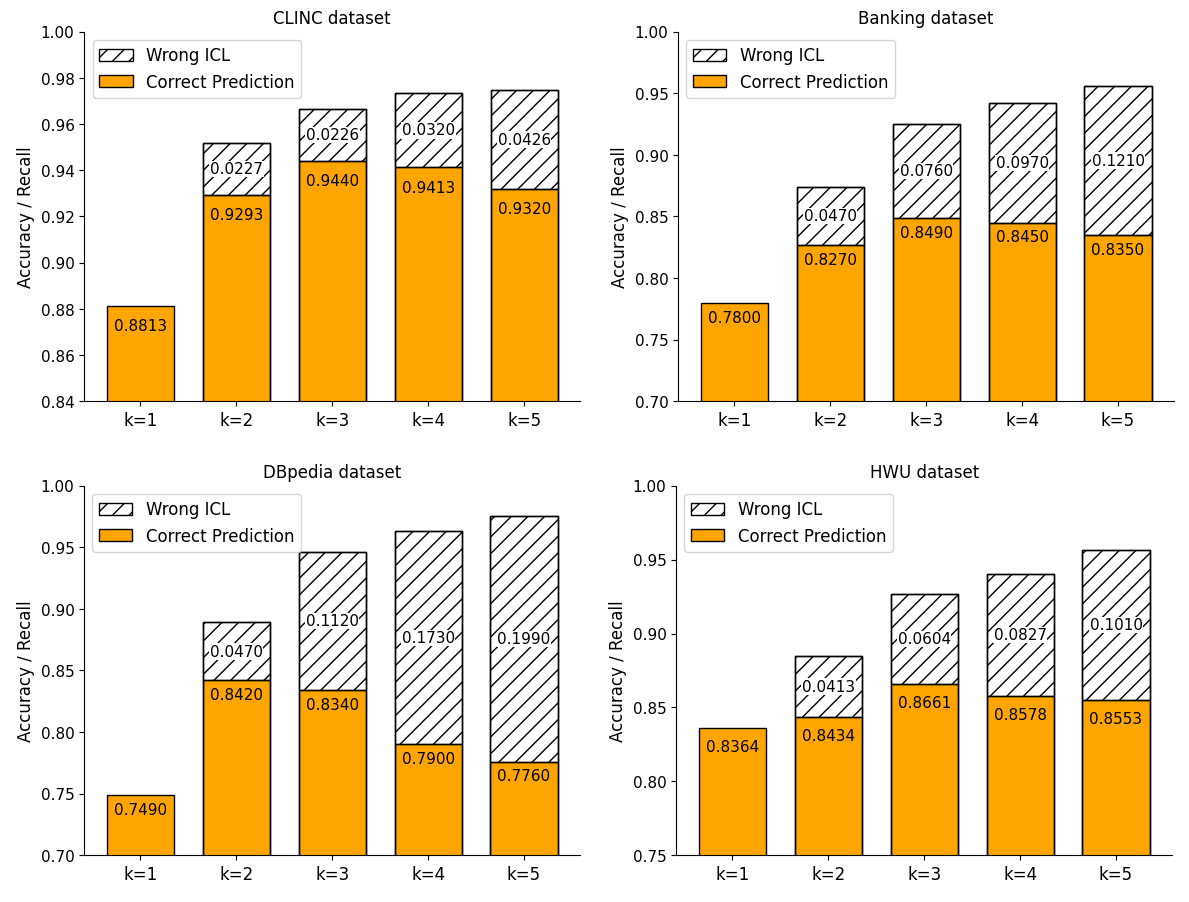}
    \caption{Accuracy of InCA and ECL recall at different $k$ values. The solid portion of each bar represents the accuracy of the model using in-context learning (ICL) with the top $k$ classes retrieved by the ECL. The leftmost column ($k$=1) represents the \textbf{accuracy of ECL alone}, where the most similar class is predicted without ICL. The dashed region indicates cases where the correct label is within the top $k$ classes retrieved by the ECL but the model's prediction is incorrect. Therefore, the total height of each bar (solid plus dashed) represents the ECL's recall of the correct classes at that $k$ value.} 
    \label{top-k}
    \vspace{-10pt}
\end{figure*}

\textbf{In-context Learning Boosts Accuracy:} The effectiveness of the in-context learning and the InCA approach becomes evident when comparing the model's final accuracy with the top-1 accuracy of the ECL alone. As shown in Figure \ref{top-k}, the final accuracy of InCA is significantly higher than the ECL's top-1 accuracy. This demonstrates the added value of in-context learning. Although the ECL alone may not be highly accurate, it effectively narrows down the relevant classes, enabling in-context learning to focus on a smaller subset and improve the overall performance.

\begin{figure}[t]
    \centering
    \includegraphics[width=\columnwidth]{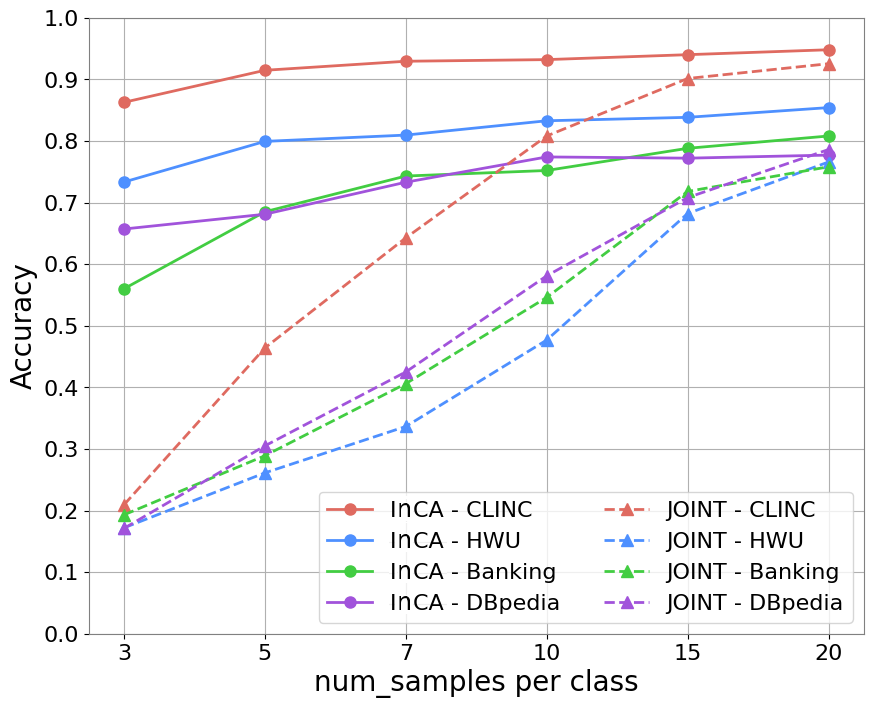}
    \caption{Performance comparison of InCA and JOINT fine-tuning across different data sizes. InCA demonstrates robust performance with limited data, particularly excelling over the fine-tuned model in data-constrained situations.} 
    \label{limited-data}
    \vspace{-12pt}
\end{figure}

\textbf{Impact of Data Size:} To assess the impact of training data size on model performance, we conducted experiments under constrained data conditions. As shown in Figure \ref{limited-data}, we compared InCA with the JOINT fine-tuning method across varying data sizes. InCA consistently maintains performance comparable to that of the full dataset, even with significantly reduced training data (e.g., 10 examples per class). Remarkably, under these limited data conditions, InCA outperforms the JOINT fine-tuning method, often considered the upper bound. These results highlight InCA's robustness and effectiveness under limited data availability.

\section{Conclusion}
\vspace{-2pt}
Existing continual learning (CL) research in NLP has primarily focused on fine-tuning or adapting LLMs for individual tasks, either by learning trainable prompts or adapters or updating the LLM's parameters. While these approaches can improve CL accuracy, their effectiveness remains limited due to catastrophic forgetting (CF). On the other hand, in-context learning with LLMs has proven highly effective across various NLP tasks. However, its application to CL is hindered by the limited context window of LLMs. As the number of tasks increases, the in-context prompt grows, often exceeding the token limit or leading to performance degradation due to overextended context, which may include irrelevant information. This paper proposed a novel method to address these challenges by leveraging an external continual learner. Our method is replay-free and does not fine-tune or adapt the LLM, treating it solely as a black box. Experiments show that our method markedly outperforms baselines, without suffering from CF.

\section{Limitations and Future Work}
\label{sec.limitations}
\vspace{-2pt}
One limitation of this work is that experiments are conducted exclusively on text classification datasets. This focus may limit the generalizability of our model to other types of NLP tasks (e.g., dialogue generation, summarization, translation, sentiment analysis, etc), which have different data characteristics and task requirements and are not suitable for \textit{class-incremental learning} because they are not classification tasks with many classes that may be learned incrementally. Although sentiment analysis is often solved as a classification task, it has a fixed number of classes, i.e., positive, negative, and neutral.  These other tasks are more suitable for \textit{task-incremental learning} or \textit{domain-incremental learning} \cite{ke2022continualsurvey}. We believe some variations of the proposed method should apply to the other NLP tasks. Designing one general method that is suitable for multiple different NLP tasks will be an interesting future research direction. 

\section*{Acknowledgments}
This work was supported in part by four NSF grants (IIS-2229876, IIS-1910424, IIS-1838770, and CNS-2225427) and a research contract from KDDI Research.

\bibliography{custom}
\raggedbottom
\pagebreak
\appendix

\onecolumn
\section{Prompts and Examples}
In this section, we detail the prompts used in various parts of our model. The provided prompts and examples are from the CLINC dataset. For different datasets, we use slightly modified versions of the same prompts based on the task at hand (e.g., intent classification, topic classification).

\subsection{Summarization Prompt}
\label{summarization-prompt}

\lstset{
    basicstyle=\ttfamily,
    breaklines=true,
    breakatwhitespace=true,
    columns=fullflexible,
    keepspaces=true,
    showstringspaces=false,
    backgroundcolor=\color{white!95!gray},
    language=Python, 
}

\begin{tcolorbox}[colback=white!95!gray, colframe=black, title=Prompt]
\begin{lstlisting}
prompt = (
    Review the following user queries and provide a summary of the intent. Keep the summary generic and avoid referencing any named entities that appear in the queries.
    
    Queries: {user_queries}
    Summary:
)
\end{lstlisting}
\end{tcolorbox}

\begin{tcolorbox}[colback=white!95!gray, colframe=black, title=Example]
Review the following user queries and provide a summary of the intent. Keep the summary generic and avoid referencing any named entities that appear in the queries.\vspace{0.25cm}

\noindent \textbf{Queries:}
\begin{itemize}[left=0pt]
    \item how would i apply for an american express
    \item show me the application for a visa card
    \item can you apply for idaho independent card for me
    \item do you have any information on how to open a new credit card
    \item is this where i apply for a new credit card
    \item is it possible to apply for a new credit card
    \item walk me through applying for a visa card
    \item what's the procedure to apply for a new credit card
    \item i want to put in an application for a line of a credit card how do i do that
    \item does bank of america give credit cards to people like me
\end{itemize}\vspace{0.2cm}

\textbf{Summary:}\vspace{0.2cm}

The intent of these queries is to apply for a new credit card. Users are looking for information on the application process, requirements, and specific credit card types. Some users are interested in applying for a specific credit card, such as an American Express or a Visa card, while others are looking for general information on how to open a new credit card.
\end{tcolorbox}
\pagebreak

\subsection{Tag Generation Prompt}
\label{tag-generation-prompt}

\begin{tcolorbox}[colback=white!95!gray, colframe=black, title=Prompt]
\begin{lstlisting}
prompt = (
    Generate descriptive tags for the following queries. Focus on user intention, relevant entities, and keywords. Extend these tags to related, unmentioned terms that are contextually relevant.
    
    Guidelines:
    Topic: Identify user intention or subject area the query pertains to.
    Entity Recognition: Focus on recognizable entities common in similar queries.
    Keywords: Extract specific terms or verbs that define the query's intent.
    Related Tags: Include tags that are related to user intention, even if not directly mentioned, to provide broader contextual understanding.
    
    Examples: {example_section}
    
    Query: {user_query}
    Tags:
)
\end{lstlisting}
\end{tcolorbox}

\begin{tcolorbox}[colback=white!95!gray, colframe=black, title=Example]
Generate descriptive tags for the following queries. Focus on user intention, relevant entities, and keywords. Extend these tags to related, unmentioned terms that are contextually relevant.\vspace{0.3cm}

\textbf{Guidelines:}\vspace{0.15cm}

Topic: Identify user intention or subject area the query pertains to.\vspace{0.1cm}

Entity Recognition: Focus on recognizable entities common in similar queries.\vspace{0.1cm}

Keywords: Extract specific terms or verbs that define the query's intent.\vspace{0.1cm}

Related Tags: Include tags that are related to user intention, even if not directly mentioned, to provide broader contextual understanding.\vspace{0.3cm}

\textbf{Examples:}\vspace{0.15cm}

Query: "Should I wear a coat today?"\\
Tags: weather advice, inquiry, clothing, temperature, coat, wear\vspace{0.1cm}

Query: "Book a table for two at a popular Italian restaurant downtown?"\\
Tags: dining reservation, Italian cuisine, booking, restaurant, table, request\vspace{0.1cm}

Query: "How can I send money to a foreign bank account using the app?"\\
Tags: international money transfer, send money, app, foreign bank, digital transfer\vspace{0.4cm}

Query: do i have to pay for carry-ons on delta\\
Tags: airline fees, carry-on, delta airlines, travel, pay, luggage

\end{tcolorbox}
\pagebreak

\subsection{Prediction Prompt}
\label{prediction-prompt}

\begin{tcolorbox}[colback=white!95!gray, colframe=black, title=Prompt]
\begin{lstlisting}
prompt = (
    Based on the given query, classify the user's intent into one of the following categories: {retrieved_classes}
    
    {class_summaries}
    
    Query: {user_query}
    Class: 
)
\end{lstlisting}
\end{tcolorbox}

\begin{tcolorbox}[colback=white!95!gray, colframe=black, title=Example]
Based on the given query, classify the user's intent into one of the following categories: direct\_deposit, income, payday\vspace{0.3cm}

\textbf{direct\_deposit}:\\
The users are inquiring about the process of setting up a Direct Deposit for their paychecks or bank accounts. They want to know how to arrange for their checks to deposit directly into their accounts and are looking for instructions or guidance on how to do this. Some users are specifically interested in setting up Direct Deposit at certain banks, while others are seeking general information on how Direct Deposit works.\vspace{0.15cm}

\textbf{income}: \\
The users are inquiring about their current or past income, salary, or earnings from their job. They want to know how much money they make or earned, and sometimes they want to calculate their total income. Some users are also interested in knowing the amount they bring in annually or their compensation.\vspace{0.15cm}

\textbf{payday}: \\
The users are inquiring about the timing of their next paycheck or payment. They want to know how often they are paid, when they can expect to be paid next, and when their next payment will be deposited. They are also interested in knowing the date or day on which they will receive their next check or be paid. Some users want to be informed about the date their most recent payment was made, while others want to plan for their next upcoming payment.\vspace{0.4cm}

Query: get my paycheck to direct deposit\\
Class: direct\_deposit
\end{tcolorbox}

\end{document}